\relax
\documentclass[letterpaper]{article} 
\usepackage{aaai22}  
\usepackage{times}  
\usepackage{helvet}  
\usepackage{courier}  
\usepackage[hyphens]{url}  
\usepackage{graphicx} 
\usepackage{natbib}  
\usepackage{caption} 
\DeclareCaptionStyle{ruled}{labelfont=normalfont,labelsep=colon,strut=off} 
\usepackage{algorithm}
\usepackage{algorithmic}

\usepackage{times}
\usepackage{latexsym}

\usepackage{amsthm}
\usepackage{helvet}
\usepackage{courier}
\usepackage{enumerate}
\usepackage{multirow}
\usepackage{array}
\usepackage{graphics}
\usepackage{graphicx}
\usepackage{subfigure}
\usepackage{amsmath}
\usepackage{multirow}
\usepackage{amssymb}
\usepackage{mathrsfs}
\usepackage{extarrows}
\usepackage{booktabs}

\usepackage{newfloat}
\usepackage{listings}
\floatstyle{ruled}
\newfloat{listing}{tb}{lst}{}
\floatname{listing}{Listing}
\title{Negative Sample is Negative in Its Own Way: Tailoring Negative Sentences for Image-Text Retrieval}
\author {
    Zhihao Fan\textsuperscript{\rm 1},
    Zhongyu Wei\textsuperscript{\rm 1,3}\footnote{Corresponding Author},
    Zejun Li\textsuperscript{\rm 1},
    Siyuan Wang\textsuperscript{\rm 1},
    Jianqing Fan\textsuperscript{\rm 2}
}
\affiliations {
    \textsuperscript{\rm 1} Fudan University,
    \textsuperscript{\rm 2} Princeton University, \\
    \textsuperscript{\rm 3} Research Institute of Intelligent and Complex Systems, Fudan University, China 
    \{fanzh18,zywei,zejunli20,wangsy18\}@fudan.edu.cn
}

\usepackage{bibentry}

\begin{document}
\maketitle
\begin{abstract}

Matching model is essential for Image-Text Retrieval framework. Existing research usually train the model with a triplet loss and explore various strategy to retrieve hard negative sentences in the dataset. We argue that current retrieval-based negative sample construction approach is limited in the scale of the dataset thus fail to identify negative sample of high difficulty for every image. We propose our TAiloring neGative Sentences with Discrimination and Correction (TAGS-DC) to generate synthetic sentences automatically as negative samples. TAGS-DC is composed of masking and refilling to generate synthetic negative sentences with higher difficulty. To keep the difficulty during training, we mutually improve the retrieval and generation through parameter sharing. To further utilize fine-grained semantic of mismatch in the negative sentence, we propose two auxiliary tasks, namely word discrimination and word correction to improve the training. In experiments, we verify the effectiveness of our model on MS-COCO and Flickr30K compared with current state-of-the-art models and demonstrates its robustness and faithfulness in the further analysis. Our code is available in \url{https://github.com/LibertFan/TAGS}.
\end{abstract}

\section{Introduction}
\label{introduction}

\begin{figure}[th!]
  \centering
    \subfigure[The diagram plots a triplet (image, positive sentence, negative sentence) as a dot is defined by matching score of the positive pair on the X-axis and that of the negative pair on the Y-axis. The matching scores are also computed by CLIP(ViT-B/32)~\cite{radford2021learning}.]{\includegraphics[width=0.465\textwidth]{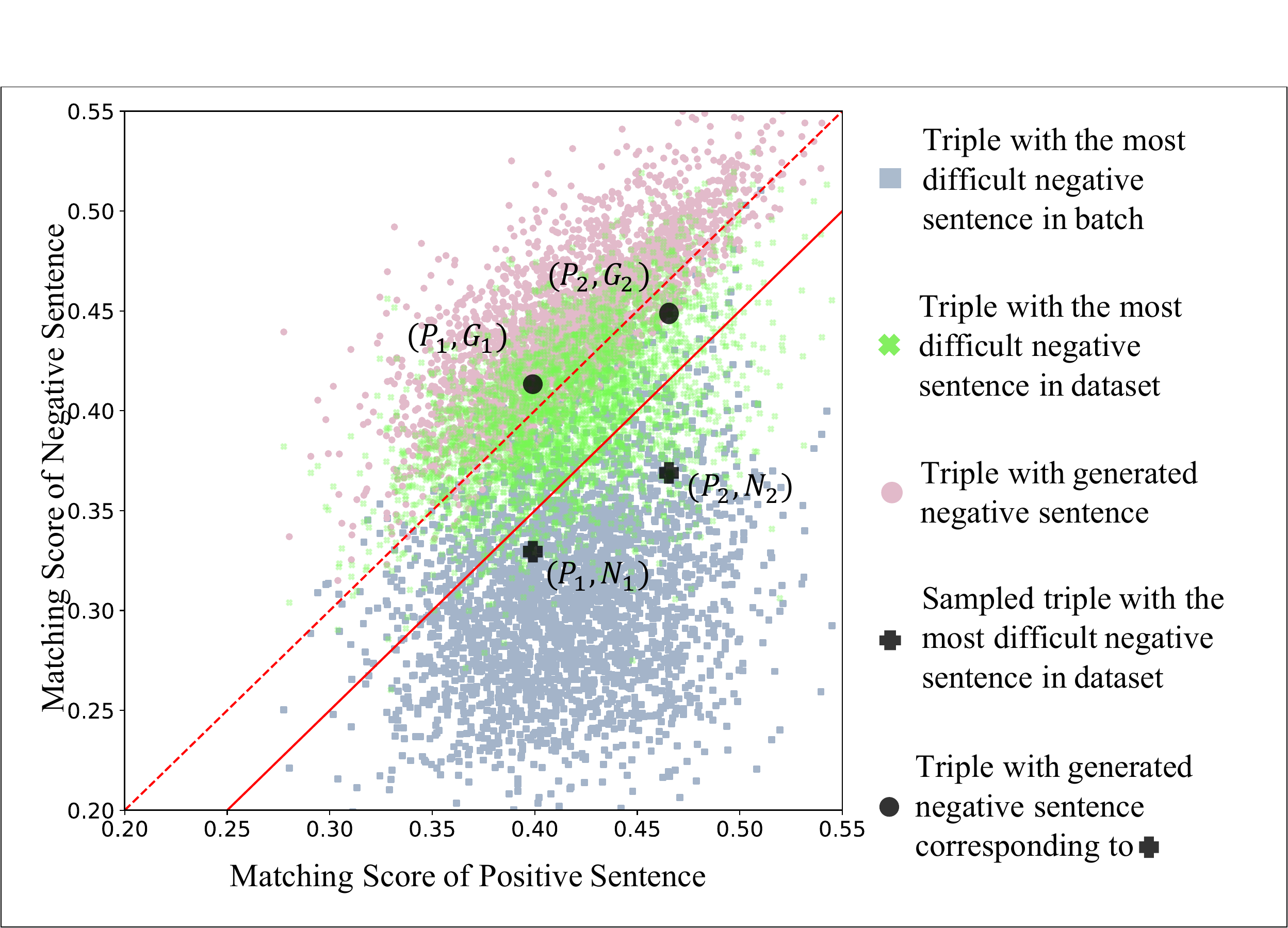}} 
    \subfigure[Two images with the positive sentence (P), the most difficult negative one (N) retrieved from dataset by CLIP and the generated negative one (G). The score is the cosine similarity computed by CLIP and larger is better. The underlined red words are non-correspondence ones to the image.]{\includegraphics[width=0.465\textwidth]{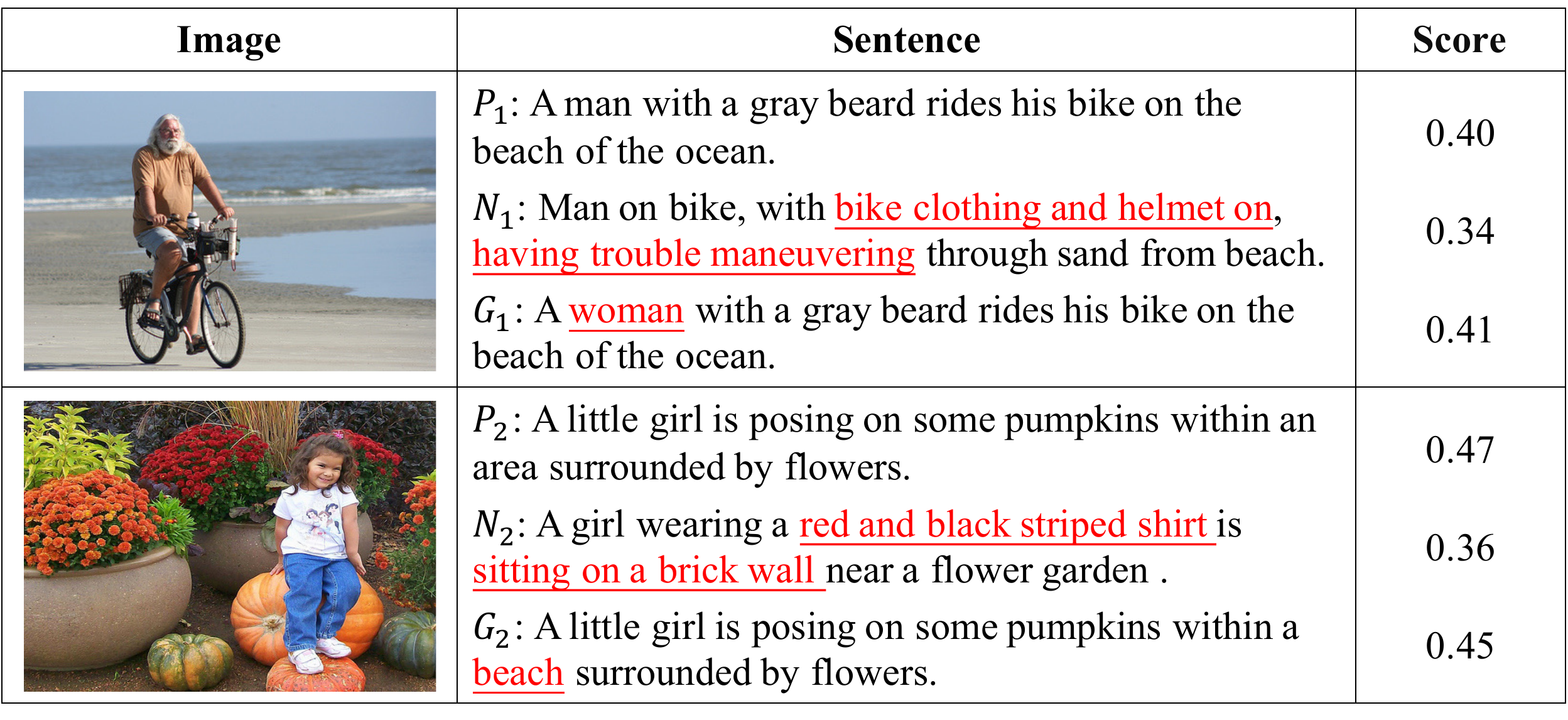}} \\
    \caption{Diagram of matching scores (a) and two examples (b) in Flickr30K~\cite{plummer2015flickr30k}.}
    \label{intro_example}
\end{figure}

The task of image-text retrieval takes a query image (sentence) as input and find out matched sentences (images) from a candidate pool. The key component of the retrieval framework is the similarity computation of an image-sentence pair and it aims to assign higher scores to positive pairs than negative ones. Triplet loss is widely applied for training. Take image-to-text as example\footnote{To keep the presentation simple and clear, we use image-to-text as example to represent tasks in both way throughout the paper.}, it constructs two image-sentence pairs using an image and two sentences (one is relevant and the other is not), and the optimization process increases the similarity of the positive pair while decreasing that of the negative one. Previous research~\cite{xuan2020hard} reveals that model trained with harder negative samples, i.e., sentences that are more difficult to be distinguished, can generally achieve better performance. In this line of work, researchers explore various strategies to search mismatched sentences for a query image, from randomly choosing mismatched sentences, to use the most similar one. The search scope moves from a single training batch~\cite{karpathy2015deep,faghri2017vse++,kiros2014unifying,socher2014grounded,lee2018stacked,li2019visual} to the whole dataset~\cite{chen2020adaptive,zhang2020learning}. Although promising results have been reported by searching for harder negative samples in a larger scope, the effectiveness is limited by the scale of the dataset. 

In order to compare the effectiveness of these strategies, we randomly sample $3,000$ images in Flickr30K~\cite{plummer2015flickr30k} and plot training triples constructed in Figure~\ref{intro_example}. Each dot stands for a triple (image, positive sentence, negative sentence), and X-axis is the matching score of the positive image-sentence pair while Y-axis is that of the negative one. In general, triples locating in the left of the dotted line are more difficult to be distinguished because matching score of the negative pair is higher than the positive one or comparable. As we can see, triples obtained by searching the most difficult mismatched sample in the batch largely locate in the right of the dotted line, and the matching scores of negative pairs are much smaller with a gap larger than 0.05 on average (in the right of the solid line). When enlarging the searching scope to the whole dataset, triples move up in positions and around 40\% of negative pairs obtain higher matching scores than positive ones. However, there are still 18\% of images can only recruit negative samples with matching score 0.05 lower than its positive counter-part. This confirms the limitation of retrieve-based negative sample construction strategy. 

To have a better understanding, we present two triples in Figure~\ref{intro_example} i.e., $(P_1, N_{1})$ and $(P_2,N_2)$ (denoted as black cross). It shows that negative sentences $N_{1}$ and $N_{2}$ describe scenes with significant differences compared with the query images, therefore, they are easy to be distinguished. Given that high percentage of images obtain these low quality of negative sentences in the dataset, we believe it is necessary to collect negative samples beyond retrieval. Instead of searching for original sentences in the dataset, we explore to construct artificial negative samples by editing positive sentences. We demonstrate two generated sentences in Figure~\ref{intro_example}, $G_{1}$ replaces ``man" with ``woman" on $P_{1}$ and $G_{2}$ replaces ``area" with ``beach" on $P_{2}$. The generated sentences obtain comparable or even higher matching scores than positive ones. We further generate artificial sentences for all images to form a new set of triples. These triples are plotted in Figure~\ref{intro_example} as pink dots. We can see all of them locate in the left side of the dotted line, which means they are more difficult to be distinguished.

In this paper, we propose TAiloring neGative Sentences (TAGS) by rewriting keywords in positive sentences of a query image to construct negative samples automatically. In specific, we employ the strategy of \emph{masking} and \emph{refilling}. In masking, we construct scene graph for the positive sentence and mask elements in the graph (objects, attributes and relations). Refilling replaces the masked original words by mismatched ones to construct the negative sample. In the training process, we further propose two word-level tasks, \emph{word discrimination} and \emph{word correction}, to incorporate fine-grained supervision into consideration. Word discrimination requires the model to distinguish which words lead to the mismatch, and word correction demands the regeneration of the original words. Both tasks evaluate the capability of the model to identify minor differences between synthetic sentences and positive ones. During inference, output of two tasks can provide fine-grained information through highlighting and revising mismatched words, and these can be regarded as the explanation for the decision made by the model to improve the interpretability. We evaluate our model on MS-COCO~\cite{lin2014microsoft} and Flickr30K~\cite{plummer2015flickr30k}. Experiment results show the effectiveness of our model.

Our contributions are three-fold: (1) We propose a generation-based method to construct negative samples to improve the training efficiency of image-text retrieval model. (2) To fully exploit the synthetic negative sentences, we propose two training tasks, word discrimination and word correction, to incorporate the fine-grained supervision to enhance the multi-modal local correspondence modeling. (3) Our model generate state-of-the-art performance on two public datasets MS-COCO and Flickr30K.

\section{Framework}

For each positive image-text pair $(I_i, T_i)$, we first generate negative sentences $\mathbb{T}_{i}^{-}$ through masking and refilling $T_i$ on the basis of masked language model (MLM) in \S\ref{section_generaion_enhanced_matching}. Second, we utilizes both retrieved and synthetic negative sentences for the training of image-text matching (IRTM and ISTM) in \S\ref{section_image_text_matching}, where synthetic negative sentences are exploited in sentence-level. Third, we propose to train the synthetic sentence generator in a dynamic way to keep pace with the upgrading of matching model. Fourth, in \S\ref{section_auxiliary_tasks}, we apply word-level tasks of word discrimination (WoD) and word correction (WoC) on $\mathbb{T}_{i}^{-}$ to discover their differences with $T_i$ for further training. MLM, IRTM, ISTM, WoC and WoD share the same backbone $M_{\theta}$ and have their own heads, namely, $\text{\emph{H}}_{\text{\emph{MLM}}}$, $\text{\emph{H}}_{\text{\emph{ITM}}}$, $\text{\emph{H}}_{\text{\emph{WoC}}}$ and $\text{\emph{H}}_{\text{\emph{WoD}}}$. The detailed training step is illustrated in Algorithm~\ref{alg:algorithm}.

\subsection{Scene-graph based Sentence Generation and Filtering}
\label{section_generaion_enhanced_matching}
In general, negative sentences with more overlapped words with positive sentences tend to obtain higher matching scores with the query image, thus are more difficult to be distinguished. Therefore, we propose to edit relevant sentences to construct negative samples for a query image. After the sentence generation, we control the quality by filtering the false negative sentences. In order to ensure the editing operates on key semantic units of the sentence, we use a strategy based on scene-graph.

\subsubsection{Scene-graph based Sentence Editing} 
The module of sentence editing takes a relevant sentence of the query image as input and output a synthetic sentence. It first identifies some key semantic units in the sentence and replace them with other words. We employ a masked language mode for this process following two steps namely, masking and refilling. 

In order to identify the key semantic of a sentence, we construct the scene-graph for a relevant sentence through scene graph parser of SPICE~\cite{anderson2016spice} following SGAE\footnote{\url{https://github.com/yangxuntu/SGAE}}~\cite{yang2019auto}. We then collect objects, relations and attributes as candidates for masking. To control the semantic offset of the synthetic sentence $T^{(k)}_{i}$, we randomly mask 15\% tokens of sentence.

In the step of refilling, we use the output head $\text{\emph{H}}_{\text{\emph{MLM}}}$, which is a two-layer feed-forward network (FFN), on top of the backbone $\text{\emph{M}}_{\theta}$ for masked language modeling. Thus, image $I_{i}$ also gets involved in MLM to guide the refilling later. The detailed computation of $ \mathcal{L}_{\text{\emph{MLM}}}$ is shown in Eq.~(\ref{masked_lm_loss}), where  $\circ$ is the function composition and $\text{\emph{NLL}}$ is the loss of negative log-likelihood.

\begin{equation}
    \begin{normalsize}
        \begin{gathered}
            \emph{MLM}: \emph{H}_{\emph{MLM}}\circ\emph{M}_{\theta}\big(I_{i},\ T^{(k)}_{i}\big)\rightarrow T_{i}/T^{(k)}_{i} \\
            \mathcal{L}_{\emph{MLM}}=\emph{NLL}\big(\emph{MLM}\big(I_{i},T^{(k)}_{i}\big), T_{i}/T^{(k)}_{i}\big) 
            \label{masked_lm_loss}
        \end{gathered}
    \end{normalsize}
\end{equation}

Then during refilling process, we put $T_{i}^{(k)}$ into $\text{MLM}$ to produce the logit scores, then sample the synthetic sentence $T_{i}^{(k,l)}$ following the distribution which originates from the logit with temperature $\tau$ as Eq.~(\ref{masked_lm_sampling}).
\begin{equation}
    \begin{normalsize}
        \begin{gathered}
            T_{i}^{(k,l)}\sim \mathop{\emph{Softmax}}\big(\emph{MLM}\big(I_{i},T_{i}^{(k)}\big)/\tau\big)
            \label{masked_lm_sampling}
        \end{gathered}
    \end{normalsize}
\end{equation}
We conduct the masking and refilling steps for $K$ and $L$ times to generate candidate synthetic sentences. 


\subsubsection{False Negative Sample Filtering}\label{criteria_c1} 
It hurts the training of using sentences that are relevant to the query image as negative samples~\cite{chuang2020debiased,huynh2020boosting}. Therefore we propose a filtering process to remove false negative ones of synthetic sentences. In vision and language datasets, each image is annotated with multiple descriptive sentences. For example, there are five in MSCOCO and Flickr30K. For a synthetic sentence, if its replaced tokens are completely included in these annotated sentences, we will treat it as relevant. Based on this, we filter synthetic sentences which are relevant.



\subsection{Image Text Matching}
\label{section_image_text_matching}
Given an image $I_{i}$ and a sentence $T_{j}$, the retrieval model assigns a matching score $s\in[0,1]$ of $(I_{i},T_{j})$ with an output head $\text{\emph{H}}_{ \text{ITM}}$, which is a one-layer FFN, as Eq.~(\ref{matching_score_target}).
\begin{equation}
    \begin{normalsize}
        \begin{gathered}
            \emph{ITM}:\emph{H}_{\emph{ITM}}\circ\emph{M}_{\theta}(I_{i},T_{j})\rightarrow s
        \end{gathered}
    \end{normalsize}
    \label{matching_score_target}
\end{equation}

Triplet loss (TripL) is widely applied in image text matching. With a hyper-parameter $\alpha$, it takes a query image (text) $U$ as an anchor for the matched (positive) image-text pair $(U,V)$ against the mismatched (negative) pair $(U,W)$ as the following equation. 
\begin{equation}
    \begin{normalsize}
        \begin{aligned}
            \emph{TripL}_{\alpha}(U,V,W)
        =\mathop{\emph{max}}\big(\alpha-\emph{ITM}(U,V)+\emph{ITM}(U,W),\ 0\big) \nonumber
        \end{aligned}
    \end{normalsize}
\end{equation}

\paragraph{Matching on Retrieved Cases} During training, for each positive image-text pair $(I_i,T_i)$, we retrieve a negative image $I^{-}_{i}$ and a sentence $T_{i}^{-}$, then employ the loss of $\text{ITM}$ in Eq.~(\ref{itm_loss}) for training,
\begin{equation}
    \begin{normalsize}
        \begin{gathered}
            \mathcal{L}_{\emph{IRTM}}=\emph{TripL}_{\alpha}\big(I_i,T_i,T_i^{-}\big)+\emph{TripL}_{\alpha}\big(T_i,I_i,I_i^{-}\big) \label{itm_loss} 
        \end{gathered}
    \end{normalsize}
\end{equation}

\paragraph{Matching on Synthetic Sentences}
First, we pick up these relatively better generated negative sentences. In practice, we compute the matching score between each synthetic negative sentence and $I_{i}$ as Eq.~(\ref{synthetic_hard_mining}), and keep a synthetic negative sentence pool $\mathbb{T}_{i}^{-}$ to make each of them as difficult as possible.
\begin{equation}
    \begin{normalsize}
        \begin{aligned}
            \mathbb{T}_{i}^{-}=\mathop{\emph{argmax}\text{-}\emph{m}}_{T^{-}_{t}\in\{T_{i}^{(k,l)}|T_{i}^{(k,l)}\ne T_{i}\}}\emph{ITM}(I_{i},T^{-}_{t})
            \label{synthetic_hard_mining}
        \end{aligned}
    \end{normalsize}
\end{equation}
where $\mathop{\emph{argmax}\text{-}\emph{m}}$ is to pick out $m$ sentences that earn the top-$m$ matching scores.

Second, with synthetic sentences $\mathbb{T}^{-}_i$ in Eq.~(\ref{synthetic_hard_mining}), we utilize them and the positive one $T_{i}$ to compute the triplet loss, and get $\mathcal{L}_{\text{\emph{ISTM}}}$ in Eq.~(\ref{synthetic_triplet_loss}).
\begin{equation}
    \begin{normalsize}
        \begin{aligned}
            \mathcal{L}_{\emph{ISTM}}=\frac{1}{|\mathbb{T}_{i}^{-}|} \sum_{T_t^{-}\in\mathbb{T}_{i}^{-}}
            \emph{TripL}_{\alpha}\big(I_i,T_i,T_t^{-}\big)
            \label{synthetic_triplet_loss}
        \end{aligned}
    \end{normalsize}
\end{equation}

\subsection{Dynamic Training Strategy of Negative Sample Generation for Image-Text Matching} The naive choice of MLM is to keep a pre-trained static one: pre-training a \text{MLM} in advance and fixing its parameters during the training of \text{ITM}. Recall that $\mathcal{L}_{\emph{ISTM}}$ encourages the ITM model to learn the pattern of synthetic sentences and keep them away from the image, we consider that negative sentences generated by the static MLM would be no longer difficult for the ITM model as the training goes on. We propose to use the dynamic MLM that shares the $\emph{M}_{\theta}$ with ITM for mutual improvement. Through the sharing, \text{MLM} continuously learns what is more relevant to the positive sentences and produce challenging negative ones for the improvement of \text{ITM}. The stronger \text{ITM} helps \text{MLM} to better identify the semantic alignment of image and keywords. MLM achieves the improvement synchronously with ITM through the interaction.

\subsection{Auxiliary Tasks to Incorporate Fine-grained Supervision}
\label{section_auxiliary_tasks}
$\mathcal{L}_{\emph{ISTM}}$ only provides sentence-level supervision and we argue it does not fully exploit the synthetic negative sentence. We introduce two auxiliary tasks to utilize the word-level difference and further enhance the model capability in multi-modal local correspondence modeling.

\begin{algorithm}[tb]
\caption{Training step of TAGS-DC}
\label{alg:algorithm}
\textbf{Input}: A positive image-text pair $(I_i, T_i)$.\\
\textbf{Parameter}: Backbone $M_{\theta}$, the head of masked language model $\text{\emph{H}}_{\text{\emph{MLM}}}$, image-text matching $\text{\emph{H}}_{\text{\emph{ITM}}}$, word discrimination $\text{\emph{H}}_{\text{\emph{WoD}}}$ and word correction $\text{\emph{H}}_{\text{\emph{WoC}}}$.
\begin{algorithmic}[1]
\STATE \emph{\# negative sentence generation}.
\STATE Initializing $\widehat{\mathbb{T}}_{i}^{-}:=\{\}$.
\FOR {$\text{\emph{k}}$ in $\text{\emph{1}},\dots,\text{\emph{K}}$}
\STATE Randomly masking $\text{\emph{T}}_{i}$ to get the masked one $\text{\emph{T}}_{i}^{(k)}$.
\STATE Computing $\mathcal{L}_{\text{\emph{MLM}}}$ in Eq.~(\ref{masked_lm_loss}) with $M_{\theta}$ and $\text{\emph{H}}_{\text{\emph{MLM}}}$.
\FOR {$\text{\emph{l}}$ in $\text{\emph{1}},\dots,\text{\emph{L}}$}
\STATE Refilling $\text{\emph{T}}_{i}^{(k)}$ to generate a synthetic sentence $\text{\emph{T}}_{i}^{(k,l)}$ following Eq.~(\ref{masked_lm_sampling}).
\IF {$\text{\emph{T}}_{i}^{(k,l)}$ satisfies criteria C1}
\STATE Adding $\text{\emph{T}}_{i}^{(k,l)}$ to $\widehat{\mathbb{T}}_{i}^{-}$  and computing its matching score with $I_{i}$.
\ENDIF
\ENDFOR
\ENDFOR
\STATE \emph{\# image text matching}.
\STATE Sampling negative image $I^{-}_{i}$ and negative sentence $T^{-}_{i}$ to compute $\mathcal{L}_{\text{\emph{IRTM}}}$ in Eq.~(\ref{itm_loss}) with $M_{\theta}$ and $\text{\emph{H}}_{\text{\emph{ITM}}}$.
\STATE Picking out top-$m$ synthetic sentences from $\widehat{\mathbb{T}}_{i}^{-}$ by the matching scores to constitute $\mathbb{T}_{i}^{-}$.
\STATE Utilizing $\mathbb{T}_{i}^{-}$ and $I_{i}$ to compute $\mathcal{L}_{\text{\emph{ISTM}}}$ in Eq.~(\ref{synthetic_triplet_loss}) with $M_{\theta}$ and $\text{\emph{H}}_{\text{\emph{ITM}}}$.
\STATE \emph{\# word discrimination and word correction}.
\FOR {$T_{t}^{-}$ in $\mathbb{T}_{i}^{-}$}
\STATE Utilizing $T_{t}^{-}$ and $I_{i}$ to compute $\mathcal{L}_{\text{\emph{WoD}}}$ in Eq.~(\ref{td_objective}) with $M_{\theta}$ and $\text{\emph{H}}_{\text{\emph{WoD}}}$.
\STATE Utilizing $T_{t}^{-}$ and $I_{i}$ to compute $\mathcal{L}_{\text{\emph{WoC}}}$ in Eq.~(\ref{tc_objective}) with $M_{\theta}$ and $\text{\emph{H}}_{\text{\emph{WoC}}}$.
\ENDFOR
\end{algorithmic}
\end{algorithm}

\begin{table*}[ht]
\begin{center}
\resizebox{\textwidth}{!}{
\begin{tabular}{lcccccccccccccc}
\midrule[1.0pt]
&\multicolumn{7}{c}{MS-COCO} &\multicolumn{7}{c}{Flickr30K}\\
\midrule[1.0pt]
\multirow{2}{*}{Model} 
&\multicolumn{3}{c}{Image-to-Text} &\multicolumn{3}{c}{Text-to-Image} & &\multicolumn{3}{c}{Image-to-Text} &\multicolumn{3}{c}{Text-to-Image} & \\
&R@1 &R@5 &R@10 &R@1 &R@5 &R@10 & RSum &R@1 &R@5 &R@10 &R@1 &R@5 &R@10 & RSum\\
\midrule[1.0pt]

\emph{SCAN} &50.4 &82.2 &90.0 &38.6 &69.3 &80.4 &410.9 &67.4 &90.3 &95.8 &48.6 &77.7 &85.2 &465.0 \\
\emph{VSRN} &53.0 &81.1 &89.4 &40.5 &70.6 &81.1 &415.7 &70.4 &89.2 &93.7 &53.0 &77.9 &85.7 &469.9 \\
\emph{MMCA} &54.0 &82.5 &90.7 &38.7 &69.7 &80.8 &416.4 &74.2 &92.8 &96.4 &54.8 &81.4 &87.8 &487.4 \\
\emph{AOQ} &55.1 &83.3 &90.8 &41.1 &71.5 &82.0 &423.8 &72.8 &91.8 &95.8 &55.3 &82.2 &88.4 &486.3 \\

\midrule[0.5pt]
\emph{UNITER+DG} & 51.4 &78.7 &87.0 &39.1 &68.0 &78.3 &402.5 &78.2 &93.0 &95.9 &66.4 &88.2 &92.2 &513.9 \\
\emph{Unicoder-VL} &62.3 &87.1 &92.8 &46.7 &76.0 &85.3 &450.2  &86.2 &96.3 &99.0 &71.5 &90.9 &94.9  &538.8 \\
\emph{LightningDOT(B)}  &64.6 &87.6 &93.5 &50.3 &78.7 &87.5 &462.2 &86.5 &97.5 &98.9 &72.6 &93.1 &96.1 &544.7 \\
\emph{ERNIE-ViL(B)} &- &- &- &- &- &-  &- &86.7 &97.8 &99.1 &\textbf{75.1} &\textbf{93.4} &96.3 &548.4 \\
\emph{UNITER(B)} &64.4 &87.4 &93.1 &50.3 &78.5 &87.2 &460.9 &85.9 &97.1 &98.8 &72.5 &92.3 &96.1 &542.7 \\
\midrule[0.5pt]
\emph{TAGS-DC(B)} &\textbf{66.6} &\textbf{88.6} &\textbf{94.0} &\textbf{51.6} &\textbf{79.1} &\textbf{87.5} &\textbf{467.4} &\textbf{87.9} &\textbf{98.1} &\textbf{99.3} &74.5 &93.3 &\textbf{96.3} &\textbf{549.4}\\
\midrule[1.0pt]
\emph{CLIP} &58.4 &81.5 &88.1 &37.8 &62.4 &72.2 &400.4 &88.0 &98.7 &\textbf{99.4} &68.7 &90.6 &95.2 &540.6 \\
\emph{LightningDOT(L)} & 65.7 &89.0 &93.7 &53.0 &\textbf{80.1} &88.0 &469.5 &87.2 &98.3 &99.0 &75.6 &94.0 &96.5 &550.6 \\
\emph{ERNIE-ViL(L)} &- &- &- &- &- &-  &- &89.2 &98.5 &99.2 &76.7 &94.1 &96.7 &554.4 \\
\emph{UNITER(L)} &65.7 &88.6 &93.8 &52.9 &79.9 &88.0 &468.9 &87.3 &98.0 &99.2 &75.6 &94.1 &96.8 &551.0 \\
\midrule[0.5pt]
\emph{TAGS-DC(L)} &\textbf{67.8} &\textbf{89.6} &\textbf{94.2} &\textbf{53.3} &80.0 &\textbf{88.0} &\textbf{472.9} &\textbf{90.6} &\textbf{98.8} &99.1 &\textbf{77.3} &\textbf{94.3} &\textbf{97.3} &\textbf{557.4} \\
\midrule[1.0pt]
\end{tabular}
}
\caption{Overall performance of the image-text retrieval. \emph{B} and \emph{L} are the base and large settings.}
\label{OverallPerformance}
\end{center}

\end{table*}

\begin{table*}[ht]
\begin{center}
\resizebox{\textwidth}{!}{
\begin{tabular}{lcccccccccccccc}
\midrule[1.0pt]
&\multicolumn{7}{c}{MS-COCO} &\multicolumn{7}{c}{Flickr30K}\\
\midrule[1.0pt]
\multirow{2}{*}{Model} 
&\multicolumn{3}{c}{Image-to-Text} &\multicolumn{3}{c}{Text-to-Image}
&
&\multicolumn{3}{c}{Image-to-Text} &\multicolumn{3}{c}{Text-to-Image}
&\\
&R@1 &R@5 &R@10 &R@1 &R@5 &R@10 &RSum &R@1 &R@5 &R@10 &R@1 &R@5 &R@10 &RSum\\
\midrule[1.0pt]

\emph{TAGS w/ WM} &64.9 &87.8 &93.3 &51.1 &78.9 &87.4 &463.4 &85.9 &97.6 &99.1 &74.2 &93.0 &96.1 &545.9\\
\emph{TAGS w/ SG} &64.1 &87.6 &93.4 &50.9 &78.8 &87.3 &462.1 &85.5 &97.4 &98.9 &73.3 &92.6 &96.0 &543.7\\
\midrule[0.5pt]
\emph{TAGS} &65.4 &88.4 &93.6 &51.3 &79.0 &87.5 &465.2 &87.2 &97.8 &99.2 &74.4 &93.1 &96.1 &547.8 \\
\midrule[1.0pt]
\end{tabular}
}
\caption{Effectiveness of Different Modules. \emph{TAGS w/ WM} means replace the scene-graph based masking with word masking in TAGS. \emph{TAGS w/ SG} means replace dynamic generator with static generator in TAGS.}
\label{analysis_result}
\end{center}
\end{table*}

\paragraph{Word Discrimination} The task is to determine whether each word of the synthetic sentence $T_{t}^{-}\in\mathbb{T}_{i}^{-}$ is matched with $I_{i}$, and we regard the replaced words of $T_{t}^{-}$ as mismatched ones and others as matched ones. The target label $\emph{G}_{t}$ of $T^{-}_{t}\in\mathbb{T}_{i}^{-}$ is determined following $\emph{G}_{t,j}=1$ if $s_{i,j}=s_{t,j}$ else $0$, where $s_{i,j}$ and $s_{t,j}$ are the $j$-th token of $T_{i}$ and $T_{t}^{-}$. We set up a new output head $\emph{H}_{\emph{WoD}}$, and the objective of word discrimination is in Eq.~(\ref{td_objective}).
\begin{equation}
    \begin{normalsize}
        \begin{gathered}
        \emph{WoD}:\emph{H}_{\emph{WoD}}\circ\emph{M}_{\theta}\big(I_{i}, T_{t}^{-}\big)\rightarrow \emph{G}_{t}\\
            \mathcal{L}_{\emph{WoD}}=\emph{NLL}\big(\emph{WoD}\big(I_{i},T_{t}^{-}\big), \emph{G}_{t}\big) 
            \label{td_objective}
        \end{gathered}
    \end{normalsize}
\end{equation}

\paragraph{Word Correction} This task is to correct these mismatched words in $T^{-}_{t}$ as Eq.~(\ref{tc_objective}). The task not only requires the model to comprehensively understand the gap between the synthetic negative sentences and the original positive ones, but also word-dependency knowledge and local cross-modal alignment to fill the gap. $\text{\emph{H}}_{\emph{WoC}}$ is the output head for word correction, and the objective is shown in Eq.~(\ref{tc_objective}).
\begin{equation}
    \begin{normalsize}
        \begin{gathered}
            \emph{WoC}:\emph{H}_{ \emph{WoC}}\circ\emph{M}_{\theta}\big(I_{i},T_{t}^{-}\big)\rightarrow T_{i}\\
            \mathcal{L}_{\emph{WoC}}=\emph{NLL}\big(\emph{WoC}\big(I_{i},T_{t}^{-}\big), T_{i}\big) 
            \label{tc_objective}
        \end{gathered}
    \end{normalsize}
\end{equation}

\subsection{Overall Training}

Details of our training step are shown in Algorithm~\ref{alg:algorithm}. The overall training loss of our model has five components as Eq.~(\ref{overall_loss}) with hyperparameters $\lambda_{\emph{IRTM}}$, $\lambda_{\emph{MLM}}$, $\lambda_{\emph{ISTM}}$, $\lambda_{\emph{WoD}}$ and $\lambda_{\emph{WoC}}$.
\begin{equation}
    \begin{normalsize}
        \begin{gathered}
            \mathcal{L}=\lambda_{\emph{IRTM}}\mathcal{L}_{\emph{IRTM}}+\lambda_{\emph{MLM}}\mathcal{L}_{\emph{MLM}} +\lambda_{\emph{ISTM}}\mathcal{L}_{\emph{ISTM}} \\
            +\lambda_{\emph{WoD}}\mathcal{L}_{\emph{WoD}}+\lambda_{\emph{WoC}}\mathcal{L}_{\text{\emph{WoC}}}
            \label{overall_loss}
        \end{gathered}
    \end{normalsize}
\end{equation} 
During inference, we employ the $\text{ITM}$ to determine the matching score of the query image (text) and the candidate text (image) as Eq.~(\ref{matching_score_target}).

\section{Experiment}
\paragraph{Dataset} We evaluate our model on MS-COCO~\cite{lin2014microsoft} and Flickr30K~\cite{plummer2015flickr30k}. In MS-COCO, each image is accompanied with 5 human annotated captions. We split the dataset following~\cite{karpathy2015deep} with 113,287 images in the training set and 5,000 images in the validation and test sets, respectively. 
Flickr30K~\cite{plummer2015flickr30k} consists of 31000 images collected from the Flickr website, and every image contains 5 text descriptions. We take the same splits as in~\cite{karpathy2015deep}, with 1000 images for validation and 1000 images for testing, and the rest for training. 

\paragraph{Models for Comparison} We compare our model with some competitive approaches, including VSE++~\cite{faghri2017vse++}, SCAN~\cite{lee2018stacked}, VSRN~\cite{li2019visual}, MMCA~\cite{wei2020multi}, and AOQ~\cite{chen2020adaptive}. We also compare with methods based on vision language pre-trained models: UNITER+DG~\cite{zhang2020learning}, Unicoder-VL~\cite{li2020unicoder}, LightningDOT~\cite{sun2021lightningdot}, UNITER~\cite{chen2020uniter}, CLIP~\cite{radford2021learning} and ERNIE-ViL~\cite{yu2020ernie}. 

\paragraph{Implementation} We employ the pre-trained UNITER as our backbone. We implement both the base (B) (12-layers, 768 hidden size and 12 attention heads) and large (L) (24-layers, 1024 hidden size and 16 attention heads) settings following UNITER~\cite{chen2020uniter}. The hyperparameters are determined by grid search. More details are described in the appendix.

\paragraph{Evaluation Metrics} We report recall at K (R@K) and Rsum. R@K is the fraction of queries for which the correct item is retrieved among the closest K points to the query. RSum is the sum of R@1+R@5+R@10 in both image-to-text and text-to-image. 

\subsection{Overall Performance}
The overall result is shown in Table~\ref{OverallPerformance}. TAGS is the model trained with generated negative samples, using dynamic training strategy. TAGS-DC is our model built on top of TAGS, further trained using two auxiliary tasks. In the base setting, our model achieves the best performance in terms of all metrics except R@1 and R@5 of in text-to-image on Flickr30K. In the large setting, our model also outperforms other models across all metrics except R@5 MS-COCO text-to-image and Flickr30K image-to-image R@10. Compared with UNITER(L), our model achieves an improvement of 4.0 and 6.4 RSum points in MS-COCO and Flickr30K.




\subsection{Ablation Study}
We further demonstrate the effectiveness of different modules, namely, scene-graph based masking (denoted as PM), dynamic sentence generation (denoted as DG) and fine-grained training tasks (denoted as WoD and WoC) in Flickr30K. Original TAGS is trained with PM and DG. TAGS-DC is further trained with Wod and WoC.

\paragraph{Scene-graph VS Word based Masking}

We replace the scene-graph based masking with word-based masking (denoted as WM) to form TAGS w/ WM. Detailed results are shown in Table~\ref{analysis_result}. WM follows the original sampling method of UNITER~\cite{chen2020uniter} that randomly sample 15\% tokens to mask, and PM is introduced in \S\ref{section_generaion_enhanced_matching}. TAGS outperforms TAGS w/ WM in terms of all metrics, and this verifies the effectiveness of PM. 

\paragraph{Dynamic VS Static Generator}
\label{ITM_DG}
We replace DG with static sentence generator (denoted as SG) to form TAGS w/ SG. The difference of TAGS and TAGS w/ SG lies in that the former shares the parameters of ITM and MLM while the latter does not. Both of them are initialized with pre-trained UNITER-base and share the same hyperparameters. In detail, we set $\lambda_{\emph{MLM}}=0.1$ and $\lambda_{\emph{ISTM}}=0.001$. The static generator is fixed as a fine-tuned UNITER+MLM model. The performance of TAGS w/ SG is not so good as TAGS. This demonstrates the effectiveness of DG.

\paragraph{WoD and WoC}
In Table~\ref{analysis_result}, TAGS-DC outperforms TAGS in both MS-COCO and Flickr30K. This reveals that word discrimination and correction contribute to the performance of ITM on the basis of TAGS. 

\section{Further Analysis}

We perform some additional experiments to explore the characteristics of the model in deep. 

\subsection{Difficulty Distribution of Samples from Dynamic and Static Generator} 
\begin{figure}[ht!]
\centering
\includegraphics[width=0.4\textwidth]{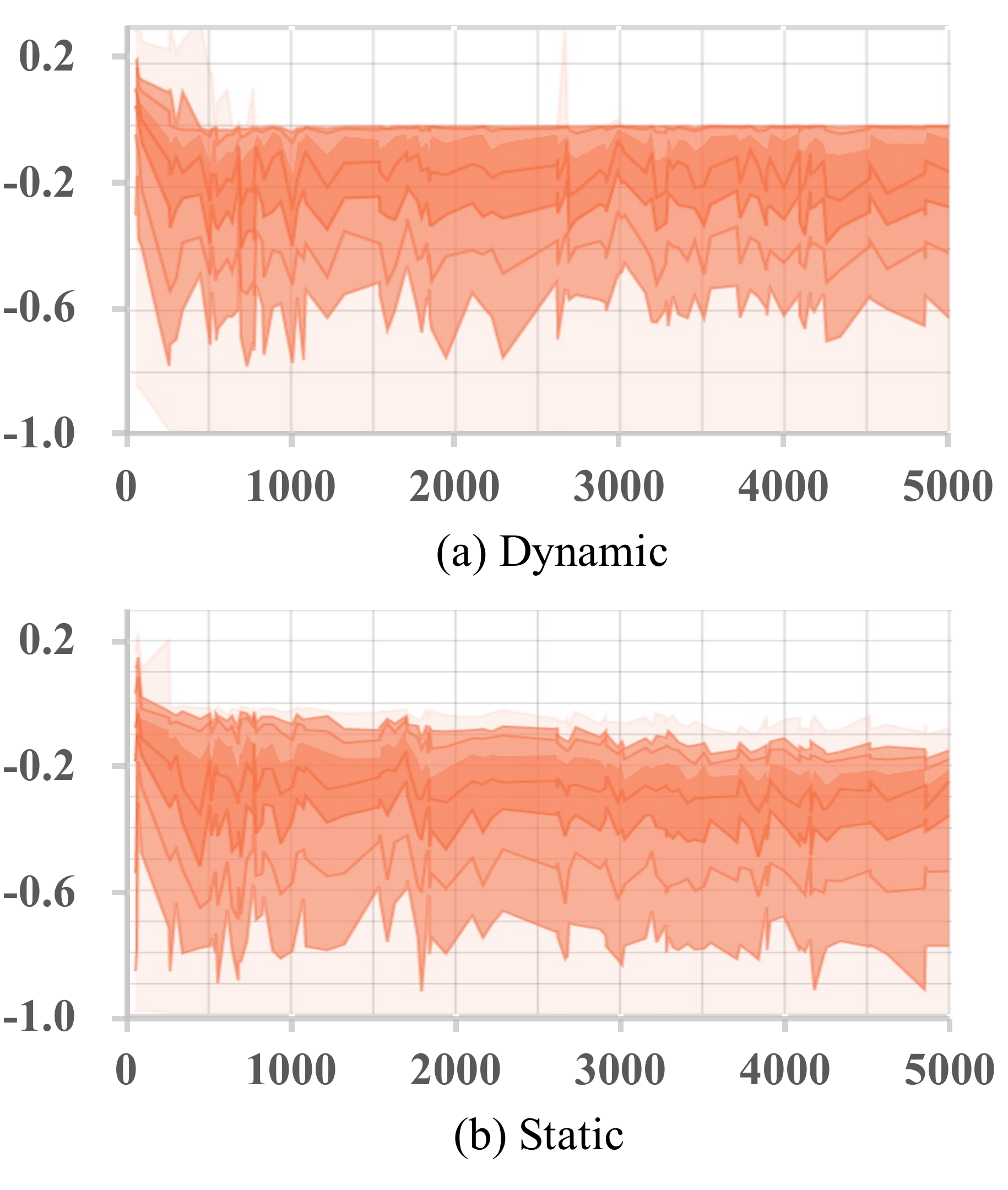}
\caption{Value $\{\text{ITM}(I_{i},T^{-}_{t})-\text{ITM}(I_{i},T_{i})\}$ distribution of triples generated by dynamic and static generators respectively during the training. X-axis is training steps.}
\label{dynamic_vs_static_sTAGS}
\end{figure}

In order to see the difficulty of negative samples constructed by various generation strategies, we plot the value distribution of samples. To evaluate the difficulty, we compute the similarity gap between the positive pair $\text{ITM}(I_{i},T_{i})$ and the negative one $\text{ITM}(I_{i},T^{-}_{t})$. We plot the value of negative pair minus positive one with respect to training steps (X-axis). In general, higher value means higher difficulty. The result is shown in Figure~\ref{dynamic_vs_static_sTAGS} where the darker color means more samples. The overall values of TAGS w/ SG (Figure~\ref{dynamic_vs_static_sTAGS} (a)) are higher than TAGS w/ DG (Figure~\ref{dynamic_vs_static_sTAGS} (b)). This implies that the static generator fails to provide negative sentences close to the image for ITM during training while our generator with dynamic generating strategy is effective.

\subsection{Quality Evaluation of Synthetic Sentences} 
We evaluate the quality of generated synthetic sentences in terms of automatic metrics and human evaluation. 
\paragraph{Fluency} We utilize the pre-trained language model GPT-2~\cite{radford2019language} to compute the perplexity of synthetic negative sentences for the measurement of their fluency. We use positive sentences in the test set of Flickr30K as original ones and generate negative samples by TAGS and VSE-C. Furthermore, we look into sentences after correction. The overall results are shown in Table~\ref{perplexity_result}. Compared with sentences produced by VSE-C, our synthetic sentences have much smaller perplexity. After correction, the fluency of synthetic sentences can be improved.

\paragraph{Human Evaluation} We perform a human evaluation to see whether all negative sentences generated are true negative. We randomly sample 200 sentences generated by TAGS and ask two annotators to determine whether the synthetic sentences are mismatched to the corresponding images. The result shows that 96.5\% of synthetic sentences generated are true negative.

\begin{table}[h]
\begin{center}
\resizebox{0.45\textwidth}{!}{
\begin{tabular}{c|c|ccc}
\midrule[1.0pt]
&Positive &Synthetic &Corrected &VSE-C \\
\midrule[0.5pt]
Perplexity &51.13 &87.63 &70.87 &292.76 \\
\midrule[1.0pt]
\end{tabular}
}
\caption{Perplexity of synthetic negative sentences.}
\label{perplexity_result}
\end{center}
\end{table}


\subsection{Negative Sentences Discrimination}
\label{evaluate_itm_with_synthetic_negative_text}
In this section, we explore to see if the generator can discriminate positive sentences from synthetic ones. We compare UNITER and TAGS. For a pair of sentences (one is positive and the other is a synthetic negative one), the generator should assign a higher score to the positive one. We report the accuracy of discrimination. We utilize two negative sentence generators TAGS and VSE-C~\cite{shi2018learning}. Two versions of TAGS with different seeds are used for cross-validation. Results are shown in Table~\ref{itm_in_MLM}. We have several findings as following. (1) TAGS2 is trained with different seed with TAGS1, but the performance of TAGS1 almost makes no difference in discriminating their generated sentences. (2) Although the synthetic sentences of VSE-C are constructed with human-efforts, TAGS also outperforms UNITER by about 9\%. (3) Three generators produce negative sentences with different distributions, but TAGS performs better than UNITER consistently. These facts validate the robustness of TAGS model.

\begin{table}[h]
\begin{center}
\resizebox{0.31\textwidth}{!}{
\begin{tabular}{c|c|ccc}
\midrule[1.0pt]
Generator &Discriminator &Accuracy \\
\midrule[1.0pt]
\multirow{2}{0.4in}{\centering \emph{TAGS1}} &\emph{TAGS1} &98.7\% \\
 &\emph{UNITER} &2.3\% \\
\midrule[1.0pt]
\multirow{2}{0.4in}{\centering \emph{TAGS2}} &\emph{TAGS1} &99.7\%  \\
 &\emph{UNITER} &2.8\% \\
\midrule[1.0pt]
\multirow{2}{0.4in}{\centering \emph{VSE-C}} &\emph{TAGS1} &96.3\%  \\
 &\emph{UNITER} &87.5\% \\
\midrule[1.0pt]
\end{tabular}
}
\caption{Accuracy of TAGS1 and UNITER in discriminating the negative sentences constructed by TAGS1, TAGS2 and VSE-C~\cite{shi2018learning}.}
\label{itm_in_MLM}
\end{center}
\end{table}

\subsection{Effectiveness of Two Auxiliary Tasks} 
\label{analysis_wod_woc}

\begin{figure}[htbp]
\centering
\includegraphics[width=0.47\textwidth]{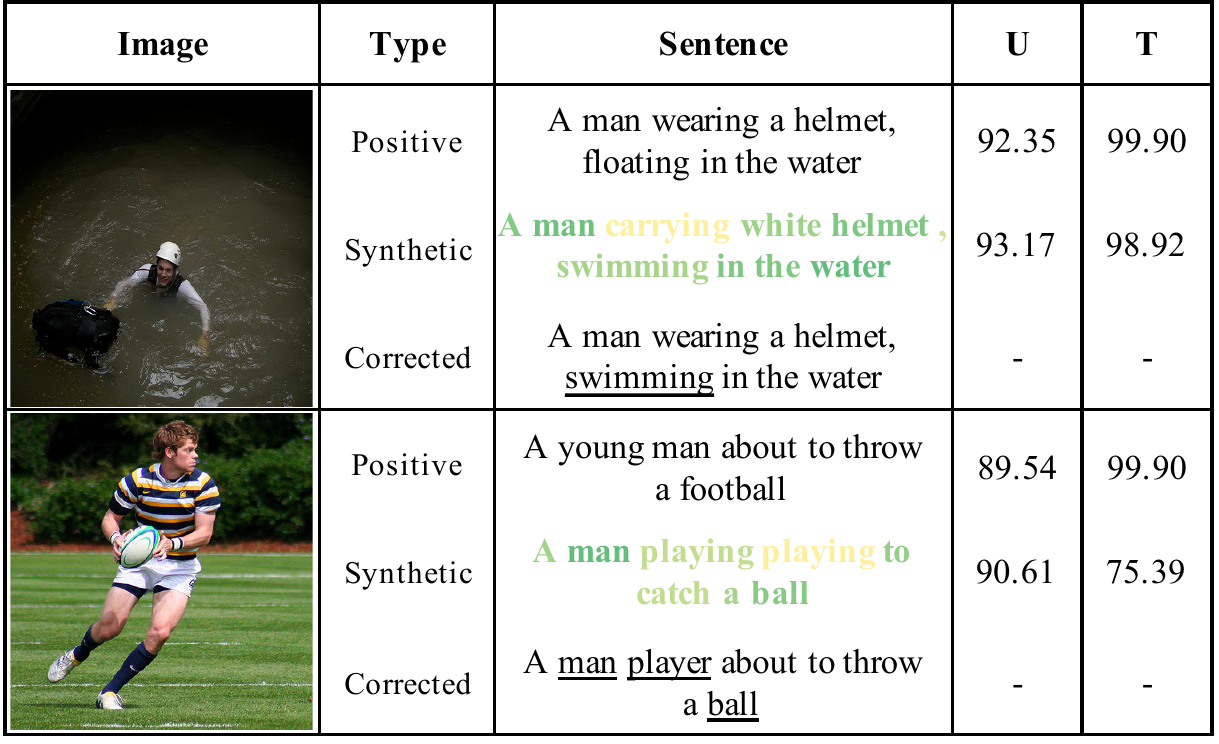}
\centering
\caption{Examples of TAGS-DC. The second column is the sentence type including positive one, synthetic one and corrected one. The third column is the corresponding sentence of the second column. The forth and fifth columns are the UNITER(U) and TAGS-DC(T) scores for the sentence in the third column, respectively. The color of word in synthetic sentences from green to yellow means the increasing of the word mismatching scores. Words with underline mean the regenerated words are different from the original ones.}
\label{case_study}
\end{figure}

We show the performance of our model in two auxiliary tasks, namely, word discrimination and correction in testing set of Flickr30K. In word discrimination, we use a threshold of 0.5 to split the positive and negative ones in terms of probability. The accuracy of word discrimination is 66.5\%. In word correction, the accuracy is 87.3\%. With the probability, we can provide additional support information accompanied to the final decision of our model. 

Two examples are presented in Figure~\ref{case_study}. (1) TAGS-DC assigns lower scores for synthetic negative sentences than positive ones, but UNITER fails. (2) Color of ``carrying'' and ``playing playing'' are yellow which means that our word discrimination successfully detect these mismatched words. Our model finds the local alignment in word-level and grammatical errors, then generates ``wearing'' and ``man player'' for correction. In examples, word discrimination marks the mismatched components and word correction provide reasons for mismatching. (3) Our model fails to identify two  mismatched words, ``swimming'', and ``ball''. Considering they are partially related to the image, our model is less effective in determining the relevance of these fuzzy words.

\section{Related Work}
\paragraph{Image-Text Retrieval} Most works in image-text retrieval focus on better feature extraction and cross-modal interaction.~\citet{nam2017dual} and~\citet{ji2019saliency} represent the image by semantics gathered from block-based attention. A line of research~\cite{lee2018stacked,wang2019camp,wang2019position,li2019visual,wang2020cross,wei2020multi,li2021unsupervised,fan2021tcic,fan2021constructing,fan2019bridging} detects features by pre-trained Faster R-CNN~\cite{ren2015faster} proposed by \citet{anderson2018bottom}. Some other methods also focus on enhancing cross-modality relationship modeling, such as the dual attention network~\cite{nam2017dual}, the stacked cross attention~\cite{lee2018stacked,liu2019focus,hu2019multi}, the graph structure attention~\cite{liu2020graph}, and the multi-modal transformer modeling~\cite{wei2020multi}. UNITER~\cite{chen2020uniter}, Unicoder~\cite{li2020unicoder} and ERNIE-ViL~\cite{yu2020ernie} follow BERT~\cite{devlin2018bert} to pre-train the vision-language transformer model on the large-scale image-text datasets, and finetune in image-text retrieval.

\paragraph{Negative Samples in Contrastive Learning} Selection strategies for negative samples have been widely studied in metric learning~\cite{schroff2015facenet,oh2016deep,Harwood_2017_ICCV,suh2019stochastic,zhang2020learning,chen2020adaptive}. \citet{wu2017sampling} employ distance weighted sampling to select more informative and stable examples. \citet{ge2018deep} present a novel hierarchical triplet loss capable of automatically collecting informative training samples via a defined hierarchical tree that encodes global context information. In the task of image-text retrieval, early works~\cite{kiros2014unifying,karpathy2015deep,socher2014grounded} utilize random negative samples for training. VSE++~\cite{faghri2017vse++} incorporates difficult negative ones in the multi-modal embedding learning. The method is widely applied in the following works~\cite{lee2018stacked,li2019visual,wei2020multi}, and achieves significant performance improvement. UNITER~\cite{chen2020uniter} randomly samples a portion of texts ($\sim$512) from dataset and pick up the hardest ones for training. AOQ~\cite{chen2020adaptive} selects these hard-to-distinguish cases from the whole dataset through a pre-trained ITM model before training and assigns hierarchical and adaptive penalties for samples with different difficulties. UNITER+DG~\cite{zhang2020learning} samples hard negative sentences according to the structure relevance based on denotation graph~\cite{plummer2015flickr30k}. These methods are retrieval-based and inspire us to find more difficult negative sentences by means of generation.~\citet{chuang2020debiased} propose a method for debiasing, i.e., correcting for the fact that some negative pairs may be false negatives. In our work, we mask keywords (objects, attributes and relationships) in positive sentence then refilling, and exclude these sentences of which each token is included in image annotated sentences. This method introduces new keywords and alleviate the generation of false negative samples.~\citet{kalantidis2020hard} consider applying mixup to produce hard negatives in latent space. In our work, we directly rewrite the positive sentences that is missing in the latent space based method, and this improves the robustness and faithfulness. The most similar work is VSE-C~\cite{shi2018learning} that attacks the VSE++~\cite{faghri2017vse++} through replacing the nouns, numerals and relations according to language priors of human and the WordNet knowledge base. Compare with VSE-C~\cite{shi2018learning}, our method has three advantages. (1) Our model does not depend on rules. (2) Our model is more flexible and can generate negative sentences with any number, but this is intractable for VSE-C. (3) The generated sentences of our model are more fluent than these of VSE-C as the results in Table~\ref{itm_in_MLM}.

\section{Conclusion}
\label{section_conclusion}
In this paper, we focus on the image-text retrieval task and find that retrieve-based negative sentence construction methods are limited by the dataset scale. To further improve the performance, we propose TAiloring neGative Sentences (TAGS). It utilizes masking and refilling to produce synthetic negative sentences as negative samples. We also set up the word discrimination and word correction to introduce word-level supervision to better exploit the synthetic negative sentences. Our model shows competitive performance in MS-COCO and Flickr30k compared with current state-of-the-art models. We also demonstrate the behavior of our model is robust and faithful. 
\bibliography{aaai22}

\end{document}